\documentclass{article}
\usepackage{authblk}
\usepackage{graphicx}
\usepackage{subcaption}
\usepackage{hyperref}
\usepackage{geometry}
\title{LEIA: Linguistic Embeddings for the Identification of Affect}
\date{}
\author[1,2]{Segun Taofeek Aroyehun}
\author[3,4]{Lukas Malik}
\author[5,3,2]{Hannah Metzler}
\author[6]{Nikolas Haimerl}
 \author[5,3,2]{Anna Di Natale}
 \author[1,2,3,5]{David Garcia \thanks{Corresponding author: \href{mailto:david.garcia@uni-konstanz.de}{david.garcia@uni-konstanz.de}}}

\affil[1]{Department of Politics and Public Administration, University of Konstanz, Konstanz, Germany}
\affil[2]{Graz University of Technology, Graz, Austria}
\affil[3]{Complexity Science Hub, Vienna, Austria}
\affil[4]{Université Paris Saclay, Paris, France}
\affil[5]{Medical University of Vienna, Vienna, Austria}
\affil[6]{Vienna University of Technology, Vienna, Austria}

\begin{document}

\maketitle

\begin{abstract}
The wealth of text data generated by social media has enabled new kinds of analysis of emotions with language models.
These models are often trained on small and costly datasets of text annotations produced by readers who guess the emotions expressed by others in social media posts. 
This affects the quality of emotion identification methods due to training data size limitations and noise in the production of labels used in model development.
We present LEIA, a model for emotion identification in text that has been trained on a dataset of more than 6 million posts with self-annotated emotion labels for happiness, affection, sadness, anger, and fear.
LEIA is based on a word masking method that enhances the learning of emotion words during model pre-training. 
LEIA achieves macro-F1 values of approximately 73 on three in-domain test datasets, outperforming other supervised and unsupervised methods in a strong benchmark that shows that LEIA generalizes across posts, users, and time periods.
We further perform an out-of-domain evaluation on five different datasets of social media and other sources, showing LEIA's robust performance across media, data collection methods, and annotation schemes.
Our results show that LEIA generalizes its classification of anger, happiness, and sadness beyond the domain it was trained on.
LEIA can be applied in future research to provide better identification of emotions in text from the perspective of the writer. The models produced for this article are publicly available at \url{https://huggingface.co/LEIA} 

\end{abstract}

\section{Introduction}

Automatic identification of emotion in text is a valuable tool to study affect through social media and other text digital traces \cite{pellert2021social}. 
Word-based methods enabled the study of mood expressions on Twitter \cite{de2012not} in relation to daylight oscillations \cite{golder2011diurnal} and of collective emotions in social resilience 
\cite{garcia2019collective}.
Rule-based methods allowed the quantification of emotion contagion on Twitter \cite{ferrara2015measuring} and the dynamics of emotions after affect labeling on social media \cite{fan2019minute}.
More advanced classification methods trained on labeled data in various languages have been used to test the effect of air pollution on happiness in Weibo posts \cite{zheng2019air}, to study the expression of emotions on Twitter about Black Lives Matter \cite{field2022analysis}, and to validate social media emotion macroscopes against survey data  \cite{pellert2022validating,garcia2021social}.
Beyond research, emotion detection from social media text has clinical potential to identify users at mental health risk \cite{calvo2017natural} and can help platforms to detect abusive language \cite{rajamanickam-etal-2020-joint}.

Despite its potential, the use of emotion detection from social media text faces important challenges. Dictionary methods applied to social media text provide user-level metrics that are weakly correlated with answers to affective questionnaires \cite{beasley2015emotional}. Furthermore, dictionary-based emotion analysis methods have weak correlations with population-level emotion prevalence  \cite{jaidka2020estimating}, but the same study shows that more advanced supervised methods bear promise to capture well-being. One of the sources of problems with the application of social media text to study emotions is the sensitivity of methods to particular domains. For example, \cite{ribeiro2016sentibench} applied out-of-the-box sentiment analysis in a benchmark of different domains and found how methods are very sensitive to the medium and text source. This is part of a general problem in which language model performance degrades with distribution shifts \cite{elsahar-galle-2019-annotate}, weakening the validity of emotion detection from text in out-of-domain (OOD) settings.

A source of error in emotion detection in social media is the way in which training labels are produced. While the target of applications is often to infer a subjective emotional state of the author of a social media post, the labels of training data are frequently produced by readers and not the authors of the post. The use of crowdsourcing can contribute to this problem, which can be alleviated by gathering several annotations per text but always carrying the potential noise source of readers not understanding the emotional state of writers. For example, a comparison between reader and writer annotations shows that they disagree 25\% of the time \cite{troiano2019crowdsourcing}.  To avoid this problem, experience sampling can be used to generate self-annotated emotion labels. For example, \cite{elayan2020stresscapes} gathered anxiety scores at the time when individuals posted tweets and compared self-reported anxiety with emotion text analysis. The results are correlations of at most 0.24, calling for studies that can leverage large datasets to identify emotional states more accurately. 

New platforms to share emotional experiences with other users offer the possibility to gather large-scale datasets with emotion self-annotations. Vent is an example that offers a particularly good source of self-annotated data, as the dataset available for researchers has millions of posts \cite{lykousas2019sharing} and the design of the platform is precisely to share emotions rather than a smaller functionality as in other platforms. Recent research on Vent has shown the difficulty to predict Vent precise mood labels from text \cite{alvarez-gonzalez-etal-2021-uncovering-limits}, but it is still left to explore how Vent can be used to infer more coarse emotion labels that can match discrete emotion classes from psychological research. In this work, we focus on a subset of Vent tags that can be mapped to standard emotional states, with the goal of training a better and more robust emotion detection model that can be applied to other text sources, especially from other social media. In the following, we present the design and development of LEIA, followed by an empirical analysis in a benchmark of in-domain and out-of-domain tests. We further analyze examples of classification errors and outputs of LEIA to understand its limitations and paths for improvement.

\section{Related Work}

Emotion classification models mainly follow feature-based or neural approaches. Feature-based methods \cite{plaza2020improved} employ handcrafted features built from resources such as emotion lexica.
Neural approaches often rely on pre-trained representations such as word embeddings and contextual language models (LMs). The use of transformer-based LMs has been shown to yield state-of-the-art performance on natural language processing benchmarks. For emotion classification, recent research works have achieved better performance using pre-trained LMs \cite{barbieri-etal-2020-tweeteval,demszky2020goemotions,nguyen-etal-2020-bertweet}.

\paragraph{Learning representations for affect.} A number of existing works learn representations for affective tasks. 
DeepMoji \cite{felbo-etal-2017-using} is a neural network trained for predicting emoji in tweets using a large distant-labeled dataset considering 64 emojis as labels.
Sentiment-specific word embeddings \cite{tang-etal-2014-learning} encode sentiment information into the vector representation of words for sentiment analysis. 
Sentiment-aware language representation learning (SentiLARE) \cite{ke-etal-2020-sentilare} incorporates part-of-speech and word polarity to enhance representation learning of a contextual language model for sentiment analysis tasks.
Another effective strategy in several natural language processing tasks is to pre-train transformer models on a large collection of text and then fine-tune the model for other downstream tasks \cite{gururangan-etal-2020-dont}, including tasks in the social media domain \cite{nguyen-etal-2020-bertweet,barbieri-etal-2020-tweeteval}. In this strategy, the adaptation step often relies on the masked language modeling objective where random tokens are masked and the model is trained to predict the masked tokens. Alternative masking strategies have been proposed to improve the pre-training task either by masking important words \cite{levine2021pmimasking} or masking words relevant for a given downstream task. 
Recently, emotion masked language modeling (eMLM) was proposed in \cite{sosea-caragea-2021-emlm} to preferentially mask emotion words for contextual language representation learning. Similar to SentiLARE, eMLM also relied on existing lexical resources by masking emotional words more frequently when training a Bidirectional Encoder Representations from Transformers (BERT) model from scratch, yielding improvements in downstream affect-related tasks. Motivated by these results, we employ eMLM in the design of LEIA as we explain below.

\paragraph{Fine-tuning strategies and model generalization} Supervised models can show a performance drop when faced with domain shifts, i.e. when they are applied to text from a domain that is not the same as the domain of their training data \cite{elsahar-galle-2019-annotate}. A recent result in computer vision \cite{kumar2022finetuning} showed that this performance gap across domains can be mitigated with a fine-tuning strategy that first performs linear probing to align the features of the prediction head with the pre-trained base model and then fine-tuning all model parameters. This approach is similar to those proposed in \cite{howard-ruder-2018-universal} and provides a further theoretical basis as well as empirical validation. Linear probing is a non-destructive and computationally cheap approach that freezes the parameters of the base model and only updates the parameters of the prediction head during training.
In this work, we consider this strategy in the context of text classification for the identification of emotion. 

\paragraph{Emotion classification datasets} 
Supervised models are trained and evaluated against emotion text datasets that are either constructed by manual labeling or automatically by using additional data sources and structures. Manually-labeled datasets are usually comparatively small while automatically-constructed datasets are built by identifying emotion-bearing patterns of expression such as hashtags in the case of Twitter. The annotation of emotion datasets can also be divided into reader-labeled and writer-labeled datasets. Reader-labeled datasets are assigned labels by the annotators post-hoc based on their perception of the emotions expressed by a given content. On the other hand, writer-labeled datasets are usually self-annotated by the writer of the message to reflect their emotion.

Most of the existing work on emotion classification has drawn on manually annotated, automatically constructed, and reader-labeled datasets.
Recently, large-scale writer-labeled datasets have been introduced \cite{lykousas2019sharing,lamprinidis2021universal} and they are yet to become part of the benchmarks of emotion detection tasks. A notable example is the Vent dataset\cite{lykousas2019sharing}, which is produced by a specialized social media platform with the goal of encouraging people to write about their feelings and provide a tag. The quality of the self-annotated emotion data drawn from Vent was examined and led to the conclusion that the tagged emotional expressions are indicative of emotional content \cite{malko-etal-2021-demonstrating}. Furthermore, the distinction between reader-labeled and writer-labeled datasets was analyzed in \cite{alvarez-gonzalez-etal-2021-uncovering-limits} with the findings indicating that classifying the emotion labels of these datasets is a hard task when considering all available labels in the platform. As supervised methods tend to perform better than unsupervised ones and gathering manual annotations is time-consuming and expensive, this kind of self-annotated datasets offers a potential alternative beyond indirect self-annotations within the text as in Twitter hashtags.

\begin{figure*}[!htbp]
    \centering
    \includegraphics[width=0.95\textwidth]{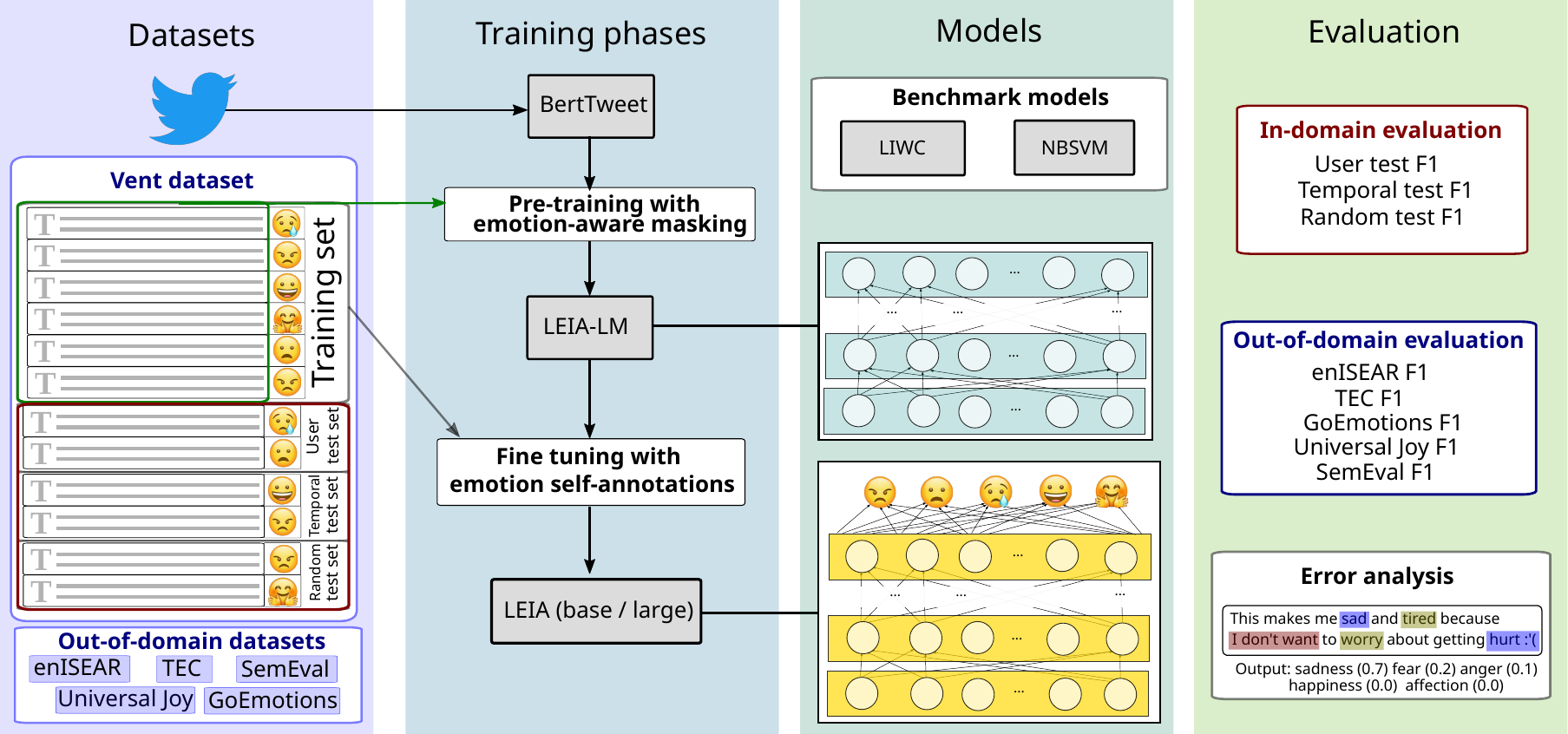}
    \caption{Overview of data sources, training steps, models, and evaluation tests.}
    \label{fig:schema}
\end{figure*}

\section{Experimental Setup}
\label{sec:experiment}
We illustrate our experimental setup in Figure \ref{fig:schema}. Next, we describe this setup more in detail starting with the datasets for training and evaluating our models, followed by details on the implementation of our proposed models and baselines.
\subsection{Datasets}
\paragraph{The Vent dataset}~\cite{lykousas2019sharing} consists of 33 Million posts from the Vent social media app. Each post is annotated by its author with an emotion tag as a way to express their emotional state to others. While the dataset has 705 emotion tags, many are temporary tags about seasonal events that do not express a clear emotional state and the most frequent tags are used on the vast majority of posts. Since Vent was designed to provide a nuanced expression of emotions rather than text classification, we mapped Vent emotion tags to a list of emotional states consistent with individual emotions from the affective science literature~\cite{ekman1999basic}. This way, we map emotion tags with words close in dimensional models of emotion~\cite{scherer2005emotion} into the same label, for example, mapping the tags \textit{angry} and \textit{annoyed} into the same label of \textit{Anger}. The precise mapping can be found in Table~\ref{tab:emotion_mapping}. Four of these emotion labels map to linguistic classes that have been consistently identified in emotional expression in text \cite{pennebaker2015development}: \textit{Sadness}, \textit{Anger}, \textit{Fear}, and \textit{Happiness.} We added a fifth category \textit{Affection}, which occurs more frequently than \textit{Happiness} and shows a social orientation of the expression of positive emotions on social media.

\begin{table}[!hbtp]
    \centering
    \caption{Mapping of Vent categories to emotion labels}
    \begin{tabular}{cc}
    \hline
         Label & Vent emotion tags \\ \hline
         Sadness & Lonely, Sad, Miserable \\ \hline
        Anger	& Angry, Annoyed, Frustrated, Furious \\ \hline
        Fear	& Anxious, Stressed, Afraid, Nervous, Worried \\ \hline
        Affection & Affectionate, Loving, Caring,  Adoring,\\ 
        & Cuddly, Supportive, Passionate, Infatuated \\ \hline
        Happiness & Happy, Excited \\ \hline
    \end{tabular}
    \label{tab:emotion_mapping}
\end{table}

We pre-process the Vent dataset to generate a cleaner dataset of posts in English that were labeled by their authors with one of the tags of Table~\ref{tab:emotion_mapping}. We remove non-English posts using three language identification tools
fasttext \footnote{\url{https://fasttext.cc/docs/en/language-identification.html}}, langid \footnote{\url{https://github.com/adbar/py3langid}}, and gcld3 \footnote{\url{https://github.com/google/cld3}}. For a post to be included in our analysis, at least two out of the three methods had to agree on detecting it as in English. After that, we remove duplicates and tag memes (invitations for a challenge to answer a question), following the approach in \cite{malko-etal-2021-demonstrating}. We remove posts with less than three words, excluding placeholders for links and user mentions in the word count. We also normalize the text by replacing multiple whitespaces with a single occurrence. We remove tab, new line and carriage return characters as well as  Hypertext Markup Language codes. The resulting dataset contains more than nine million posts with metadata including the emotion labels, pseudonymized user ids, and timestamps when the post was written.

\begin{table*}[!htbp]
    \centering
    \caption{Frequency of occurrence of the labels on the data splits of the Vent dataset after pre-processing. The proportion of the total number of instances within the sample is in parenthesis.}
      \begin{tabular}{c c c ccc}
        \hline
        & Train& Development & User Test & Temporal Test & Random Test \\ \hline
               
         Sadness &	1,712,985 (27\%) & 199,890 (28\%) & 262,999 (27\%) & 293,993 (30\%) & 264,906 (27\%) \\
        Anger	& 1,517,282 (24\%) & 147,778(21\%) & 224,997 (23\%) & 205,598 (21\%) & 226,068 (23\%)  \\
        Fear	& 1,341,624 (21\%) & 138,929 (20\%) & 198,264 (21\%) & 185,461 (19\%) & 201,563 (21\%)\\
        Affection &	979,019 (15\%) & 144,175 (20\%) & 161,018 (17\%) & 191,022 (20\%) & 158,017 (16\%)\\
        Happiness & 795,363 (13\%) & 74,369 (11\%) & 118,290 (12\%) & 91,127 (9\%) & 116,647 (12\%)\\ \hline
         \textbf{Total}& 6,346,273 &  705,141 & 965,568 & 967,201 &  967,201 \\ \hline
    \end{tabular}
    \label{tab:label_freq}
\end{table*}

\paragraph{In-domain evaluation datasets}
An overview of this study can be seen in Figure \ref{fig:schema}, including data sources and data splits for in-domain evaluation. We split the pre-processed Vent dataset into a training/development/test split with three disjoint test datasets to assess the capability of the model to generalize emotion identification. The random test set contains a uniformly random selection of 10\% of all posts in the Vent dataset. The user test set consists of all posts written by a random sample of 10\% of the users. This way, no post in the training set has been written by any of the users in the user test set. The temporal test set contains the last 10\% of the posts according to their timestamp, thus allowing us to evaluate the model with future data with respect to its training set. We additionally extracted another 10\% random set from the remaining posts as a development set to guide model design before the final run of all tests. All these subsets are disjoint and the three tests allow us to evaluate if and how the model generalizes across posts, users, and time.  The resulting exact counts of posts and emotion labels in all splits can be found in Table~\ref{tab:label_freq}.

\paragraph{Out-of-domain evaluation datasets}
To evaluate if models learn about emotional expression beyond the domain of Vent as a social platform, we include five OOD datasets with emotion labels and texts associated with the emotions. The OOD datasets are the following:

\begin{itemize}
  \item enISEAR  \cite{troiano2019crowdsourcing} is a dataset of emotional event descriptions in English using the International Survey
on Emotion Antecedents and Reactions (ISEAR) approach \cite{scherer1994evidence} via crowdsourcing. Annotators generated event-focused emotion descriptions using the template: ``\textit{I felt [emotion] when/because [situation]}''. While the study included annotations by readers, we only use the annotation of the author of the text to evaluate models. The dataset consists of 1001 instances for seven emotions, four of which match our emotion labels to provide an out-of-domain test. We design the task as a prediction of the text in which we have replaced the emotion word with the placeholder \textit{mask}, which is a special token common in language models to denote a missing word. enISEAR is generated by asking participants to describe an emotion-inducing situation, a design that limits its external validity with respect to social media  but that has the highest standard of internal validity with text annotations produced in a controlled setup. We consider enISEAR as the out-of-domain dataset most relevant to test the psychological validity of the emotion detection of models, while other datasets from social media are necessary to evaluate models in other domains once this psychological validity level is clear.
  
    \item GoEmotions \cite{demszky2020goemotions} is a corpus of English comments extracted from Reddit with manual annotations for multiple emotions.
    It is a reader-labeled emotion dataset with labels assigned when at least three annotators gave the same label to a comment. For our out-of-domain test, we include the subset of the test split with a single label from among the Ekman category of the dataset, thus having \textit{Sadness}, \textit{Anger}, \textit{Fear}, and \textit{Joy} as a general positive emotion label.
    
    \item TEC \cite{mohammad2012emotional} is a corpus of tweets posted between Nov. 15, 2011 and Dec. 6, 2011 with self-label for emotions using emotion-word hashtags. The hashtags serve as the emotion label for classification and are removed from the tweet texts. We sample 10\% of the dataset at random as our out-of-domain test set. Since the hashtags are assigned by the authors of the tweets, the dataset can be considered labeled from the perspective of the writer.

    \item Universal Joy \cite{lamprinidis2021universal} is a collection of anonymized public Facebook posts in 18 languages labeled with five emotions: anger, anticipation, fear, joy, and sadness. The labels are derived from the Facebook ``feelings tag'' provided by the writers of the posts. We use the English subset of the test set for our analysis.
    
    \item SemEval \cite{mohammad-etal-2018-semeval} is a collection of tweets in three languages from 2016 and 2017 collected from Twitter using emotion keywords as queries. Subsequently, matching tweets were annotated by crowdworkers for emotion intensity, valence, and basic emotion classes. This dataset was the benchmark data for the competition about affect detection in SemEval. Here, we use the test data by including only instances with a single label that correspond to one of the labels in our model.
    
\end{itemize}

Note that for the OOD datasets (GoEmotions, TEC, Universal Joy, and SemEval), we use only the test sample for OOD evaluation and exclude other training or development samples. We do this to provide an evaluation that can be compared to previous and future supervised methods that use the training samples.

\begin{table}[!htbp]
    \centering
    \caption{Frequency of occurrence of the labels on the test sets of out-of-domain datasets }
        \begin{tabular}{c|cccc|c}
        \hline
        Dataset    & Sadness & Anger & Fear & Happiness & \textbf{Total} \\ \hline
        enISEAR & 143 & 143 & 143 & 143 & 572 \\
        TEC & 765 & 305 & 499 &   1,627 &  3,196 \\
        GoEmotions & 259 & 520 & 77 &  1,598 &   2,454 \\
        Universal Joy & 128 & 58 & 11 & 384 & 581 \\
        SemEval & 312 & 511 & 165 &  706 & 1,694 \\
        \hline
    \end{tabular}    
    
    \label{tab:label_freq2}
\end{table}

Based on our selection criteria, we find only 11 tweets with the Affection label in the SemEval dataset.
So, we consider \textit{Happiness} and \textit{Affection} to be the \textit{Happiness} emotion label, which limits the nuance in which we can assess classifications within positive emotions in out-of-domain settings but still enables a wider differentiation between general positive emotions and three negative emotions. Descriptive statistics of the counts and proportions of labels in the five datasets can be found in Table~\ref{tab:label_freq2}.

We use the in-domain and OOD datasets to evaluate the performance of models in our experimental setup. 
We calculate the macro-averaged F1 score over all emotion labels and report results with the F1 score of each of the emotion labels, as their frequencies greatly differ in several of the datasets we use for evaluation.

\subsection{Models}
\label{model}
\paragraph{Model design and pre-training} 
Pre-trained language models have shown state-of-the-art performance on many natural language processing tasks. We expect language models pre-trained on social media data to perform better on the Vent dataset. In preliminary experiments using performance on the development set, we test three pre-trained models based on the Robustly optimized BERT approach (RoBERTa) architecture and pre-training: Roberta-base~\cite{liu2019roberta}, Twitter-RoBERTa~\cite{barbieri-etal-2020-tweeteval}, and BERTweet-base~\cite{nguyen-etal-2020-bertweet}. 
BERTweet-base had the best performance on the development set and thus we chose to continue our work with BERTweet-base and its large version, BERTweet-large, in all our experiments. BERTweet-base and BERTweet-large are transformers model pre-trained on 850M tweets with 12 and 24 layers, respectively.  BERTweet-base has a maximum sequence length of 128 (sub)words while BERTweet-large has a maximum sequence length of 512 (sub)words~\cite{nguyen-etal-2020-bertweet}.
Before training a classifier on the training set, we pre-train BERTweet-base (BERTweet-large) on the text of Vent posts in the training set ignoring all emotion labels. We perform task-adaptive pre-training \cite{gururangan-etal-2020-dont} by preferentially masking emotion words using eMLM.
We use the emotion terms in the NRC emotion lexicon~\cite{mohammad-turney-2010-emotions,Mohammad13} as it is one of the most extensive emotion lexicons available. We set the probability of masking emotion words to 0.5 following previous work~\cite{sosea-caragea-2021-emlm}. 
We train with the eMLM objective for 100K steps using the AdamW optimizer~\cite{loshchilov2018decoupled}, a learning rate of $5*10^{-5}$, and a batch size of 128. 
We name the resulting models LEIA-LM-base and LEIA-LM-large, i.e. the result of our pre-training of BERTweet-base and BERTweet-large respectively. On an NVIDIA RTX8000 GPU, pre-training takes approximately a week for the base model and a month for the large model. 

\paragraph{Model fine-tuning with labeled data} We implement a multiclass classifier for the five emotion labels: \textit{Anger}, \textit{Fear}, \textit{Sadness}, \textit{Happiness}, and \textit{Affection}. We train classifiers starting from LEIA-LM-base and LEIA-LM-large using a two-step approach. First, we perform linear probing to initialize the classifier head and then full fine-tuning of the model.  For linear probing, only the classifier head is randomly initialized and trained on the training dataset while the remaining model parameters are fixed. This initial step can be seen as a way to align the features of the prediction head and the base model to minimize feature distortion \cite{kumar2022finetuning}. In the subsequent full fine-tuning step, the prediction head is initialized from the parameters learned from the initial linear probing step.  We also fine-tune a BERTweet-base and a BERTweet-large model without the eMLM step. 
To improve model generalization, we average model weights~\cite{wortsman2022model} of the two model variants (one with eMLM and one without eMLM) for each of the base and large architectures. The resulting models are respectively named LEIA-base and LEIA-large. We show the performance of the intermediate model variants on the in-domain and OOD test sets as an appendix in Tables~\ref{tab:intermediate_indomain_comp} and~\ref{tab:intermediate_ood_comp}.
For the linear probing step, we use a learning rate of $5*10^{-4}$ and train only the classifier head while the other layers are frozen for 1000 steps. 
For fine-tuning, we set the learning rate to $10^{-5}$ with a constant learning rate schedule, embedding dropout of 0.1, weight decay factor of 0.01, and a label smoothing factor of 0.1. We train for 5 epochs using AdamW optimizer with an effective batch size of 256 and a maximum sequence length of 128. We jointly optimize a supervised contrastive loss and a cross-entropy loss~\cite{gunel2021supervised}. The supervised contrastive loss ensures that the model captures the similarity between examples within a class while contrasting them with examples from other classes. This approach has been shown to aid model generalization. Following prior work~\cite{gunel2021supervised}, we set the weight of the contrastive loss to 0.9 and the temperature parameter to 0.3. The fine-tuning process takes approximately 24 hours for the base-sized model and 60 hours for the large-sized model on an Nvidia RTX8000 GPU with 48GB memory.

\paragraph{Baselines} As baselines, we use the popular Linguistic Inquiry and Word Count (LIWC) dictionary approach \cite{boyddevelopment2022} and a Naive Bayes Support Vector Machine (NBSVM) as a supervised baseline. 
For the LIWC approach, we map the score for the relevant LIWC categories to emotion labels as follows: \textit{$emo\_anger$} to Anger, \textit{emo\_anx} to Fear, \textit{emo\_sad} to Sadness, and \textit{emo\_pos} to Happiness. We did not find a category that can be mapped to Affection in the LIWC categories, thus considering only 4 classes for the dictionary-based baseline.
We convert the multiclass result of LIWC to a binary classification task for each emotion label using the ``one-vs-rest'' setting. For Sadness category as an example, we consider instances within the Sadness category as having a label of 1 while all other examples are assigned a label of 0. We first compute the base frequency as the average of the LIWC score for each emotion category on the Vent development set and divide the LIWC score for each post in the test set with the base frequency. If the quotient is greater than 1, we predict that the respective emotion is present in the post otherwise absent. We opt for this option in order to be able to handle cases where the LIWC dictionaries do not match any word in a given post which results in a score of 0 across all labels.

We use NBSVM \cite{wang-manning-2012-baselines} as a supervised baseline. NBSVM is a strong baseline for text classification that uses Naive Bayes features for unigrams as input representation. We use the implementation in Ktrain \cite{maiya2022ktrain} with a vocabulary size of 64K.

\section{Results and Analysis}
In this section, we report the performance of LEIA-base and LEIA-large in both in-domain and out-of-domain scenarios. We include the macro-F1 score and bootstrapping confidence intervals obtained from 10000 bootstrap samples. We provide an error analysis on a sample of incorrect model predictions. We end by assessing the salient features on selected examples of model predictions.

\paragraph{In-domain results} Table \ref{tab:in_domain_test_scores} shows that LEIA-base and LEIA-large outperform all models in all three Vent test samples, achieving a Macro-F1 of about 73 on random posts, text from unseen users and different time periods. Model performance is comparable across all three test sets, which indicates that its F1 score is not achieved by exploiting biases of user activity or high-volume time periods. 
The dictionary approach has the lowest macro-F1 scores, being significantly outperformed by LEIA-base and LEIA-large.
The supervised approach of NBSVM achieves macro-F1 scores of about 60 but is still substantially and significantly outperformed by LEIA-base and LEIA-large.

\begin{table}[!h]
\centering
\caption{Macro-F1 scores on the Vent test sets. 95\% Confidence interval in square brackets (computed over 10000 bootstrap samples). For LIWC, we only consider 4 out of 5 labels and perform binary classification for each label using the ``one-vs-rest'' setting.}
\begin{tabular}{ccccc}
         & LIWC                   & NBSVM                  & LEIA-base              & LEIA-large             \\ \hline
         
        user & 26.49{[}26.40,26.60{]} & 60.15{[}60.05,60.25{]} & 72.92{[}72.82,73.02{]} & 73.37{[}73.28,73.46{]} \\
        temporal & 29.05{[}28.96,29.14{]} & 60.52{[}60.42,60.63{]} & 73.03{[}72.95,73.11{]} & 73.43{[}73.34,73.53{]} \\
        random   & 26.46{[}26.37,26.54{]} & 60.26{[}60.16,60.36{]} & 73.02{[}72.94,73.12{]} & 73.57{[}73.48,73.66{]} \\ \hline

\end{tabular}
\label{tab:in_domain_test_scores}
\end{table}

\begin{figure*}[!htbp]
    \centering
        \includegraphics[width=0.95\textwidth]{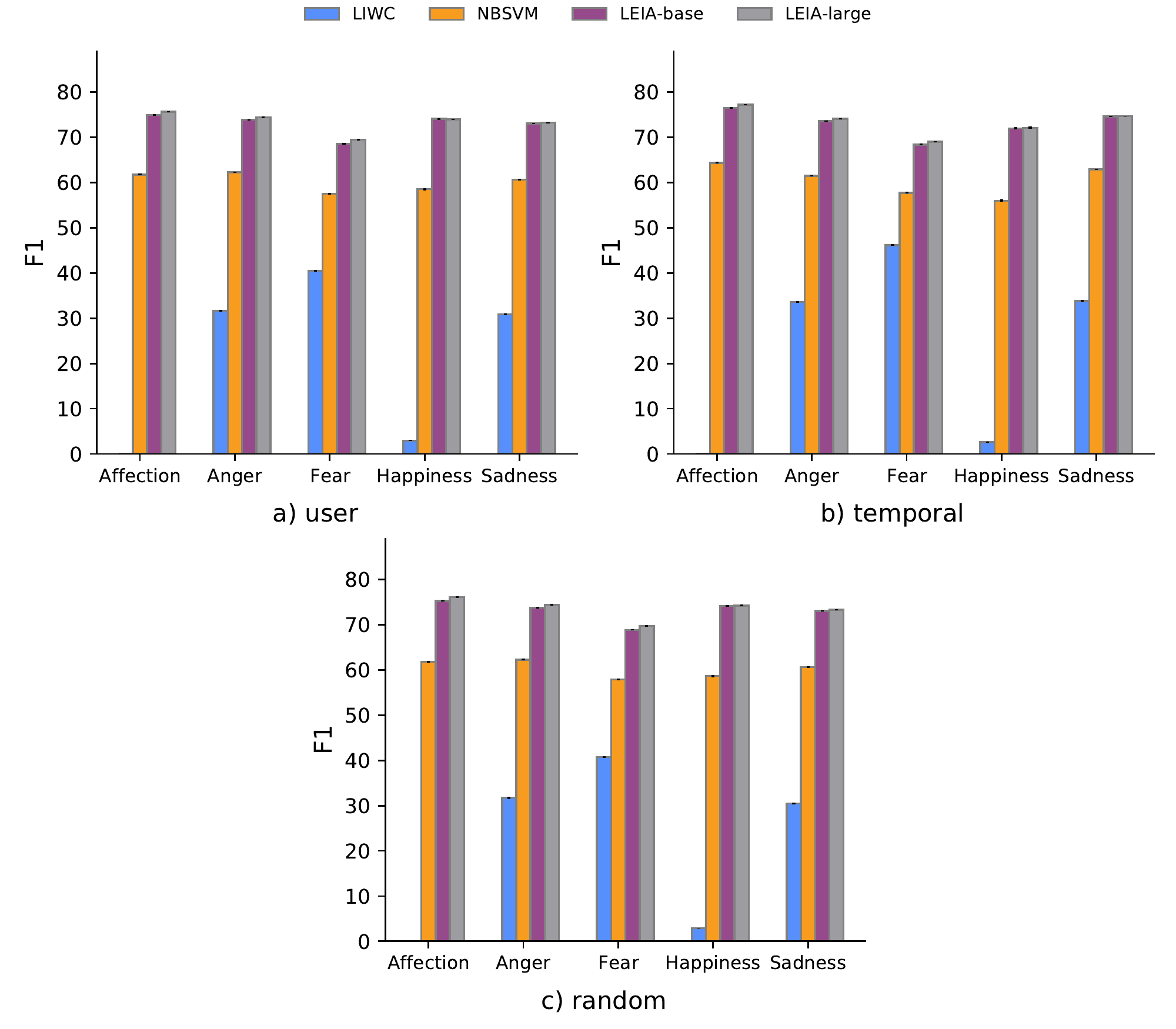}
        \caption{Results within the Vent dataset in the three test samples. Error bars show bootstrap 95\% confidence intervals and may be too small to be visible due to the large sample sizes.
        }
        \label{user}
    \label{fig:label_ID}
\end{figure*}

Figure \ref{fig:label_ID} shows a breakdown of F1 per emotion class in the in-domain test samples. LEIA-base and LEIA-large show consistently high F1 score for all emotion classes. This shows that the general performance of LEIA-base and LEIA-large is not as a result of bias from higher performance on majority class. The only class that has a slightly lower F1 is Fear, but LEIA-base and LEIA-large still outperform all other methods on it. One observation is that NBSVM also performs slightly worse for Fear than for other emotions, in contrast with LIWC, which obtains a comparatively better performance in the Fear category.

\begin{figure*}[!htbp]
        \centering
        \includegraphics[width=0.95\textwidth]{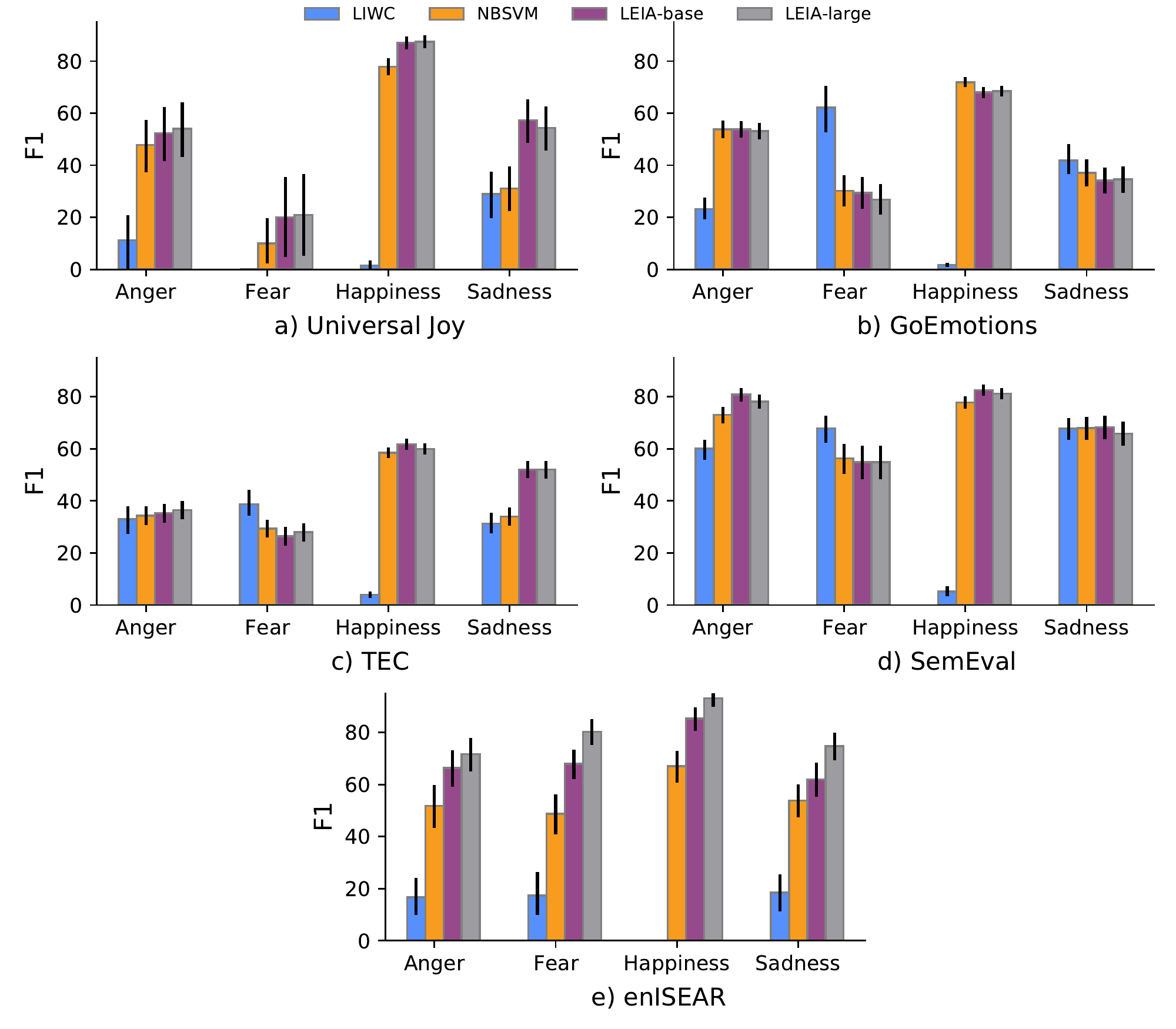}
    \caption{F1 score for each label for the out-of-domain datasets. Error bars represent confidence intervals computed using bootstrapping with replacement. Missing bars correspond to F1 of 0.
    }
    \label{fig:label_OOD}
\end{figure*}

\paragraph{Out-of-domain results}
Our out-of-domain benchmark shows that LEIA can detect emotional states in other types of text and social media platforms beyond Vent. Table \ref{tab:ood_test_scores} shows the Macro-F1 scores for the five out-of-domain test sets. LEIA-base and LEIA-large have significantly higher F1 scores than all other methods when evaluated on 4 out of the 5 OOD datasets. The NBSVM has a
comparable performance in the GoEmotions dataset, where the F1 of NBSVM and of LEIA-base are not significantly different. We also observe that a larger model does not necessarily lead to better performance on OOD datasets, as LEIA-large only shows a substantially different performance on the enISEAR dataset.
Figure \ref{fig:label_OOD} shows the F1 score for each class on the OOD datasets. 
In general, LEIA often outperforms baselines across labels. 
LEIA is significantly better than the baselines for Happiness and Sadness in the Universal Joy and TEC datasets, for all emotions in the enISEAR dataset, and for all emotions except Fear and Sadness in the SemEval dataset. 
On the GoEmotions dataset, LEIA is tied with NBSVM as the best method to detect Anger as F1 score is not significantly different. The Fear class evaluation poses some challenges in this OOD evaluation since evaluation samples for this class can be very small (e.g. 11 posts in Universal Joy and 77 in GoEmotions). In the case of Fear, the dictionary approach performs significantly better than the supervised approaches on GoEmotions, SemEval, and TEC. Recall that the dictionary approach is based on a binary classification setting which is easier than a multiclass classification setting. Despite this, the performance of the dictionary approach is significantly lower for Happiness, even reaching an F1 score of 0 on the enISEAR dataset. This trend is similar to the performance observed on the in-domain test sets.

\begin{table*}[!h]
  \centering
  \caption{Macro-F1 scores on out-of-domain datasets. 95\% Confidence intervals in square brackets (computed over 10000 bootstrap samples).}
\begin{tabular}{ccccc}
       & LIWC                   & NBSVM                  & LEIA-base              & LEIA-large             \\ \hline
       
Universal Joy  & 10.43[7.01,13.60] & 41.70{[}37.36,46.08{]}  & 54.18{[}48.79,59.88{]} & 54.17{[}48.68,59.84{]} \\
GoEmotions & 32.23[28.87,35.18] & 48.23{[}45.85,50.59{]} & 46.31{[}43.98,48.72{]} & 45.75{[}43.45,48.09{]} \\
TEC        & 26.75[24.84,28.61] & 39.07{[}37.28,40.92{]} & 43.87{[}42.05,45.61{]} & 44.12{[}42.34,45.89{]} \\
SemEval    & 50.21[48.55,51.92] & 68.77{[}66.29,71.25{]} & 71.68{[}69.18,74.19{]} & 70.04{[}67.48,72.52{]} \\
enISEAR    & 13.11[9.81,16.41] & 55.33{[}51.22,59.41{]} & 70.37{[}66.63,74.01{]} & 79.94{[}76.69,83.14{]} \\ \hline
\end{tabular}
\label{tab:ood_test_scores}
\end{table*}

We can conclude that LEIA shows a good generalization beyond the domain it was trained on, first by achieving very high performance in enISEAR, the test closest to psychological methodology, but also achieving good performance for datasets that include posts from other social media such as Twitter and Facebook. The lower performance recorded for Fear on the out-of-domain test sets is not surprising as the model performance on this category tends to be lower on the in-domain test sets too. LEIA achieves a consistently high score for Happiness on the out-of-domain test sets despite the fact that it is one of the least frequent categories in the training set. This suggests that it constitutes an easier category for the model to recognize across domains than more nuanced negative emotions.


\paragraph{Error analysis} We examine a random sample of 50 incorrect predictions from the user test split (10 per label) of the Vent dataset.
We find that majority of errors in the sample can be categorized into the following cases:

\begin{enumerate}
\item Messages conveying an expectation of a positive outcome while the self-assigned label has negative valence (e.g., I need a good online game). These cases represent situations where the text is very similar to positive texts but subtle signals point toward negative states.
\item Expressions of both positive and negative emotions at the same time. These are assigned a single label by design but other labelling schemes could cope with mixed emotions.
\item Use of figurative expressions such as humor or sarcasm that the model does not recognize. 
\item Very short posts that do not contain indications about the emotional state of the author (e.g., going for a coffee) where additional context is required. 
\item Few instances where we find the model prediction more plausible than the assigned label.

\end{enumerate}

\begin{figure*}[h!]
    \centering
    \includegraphics[width=0.95\textwidth]{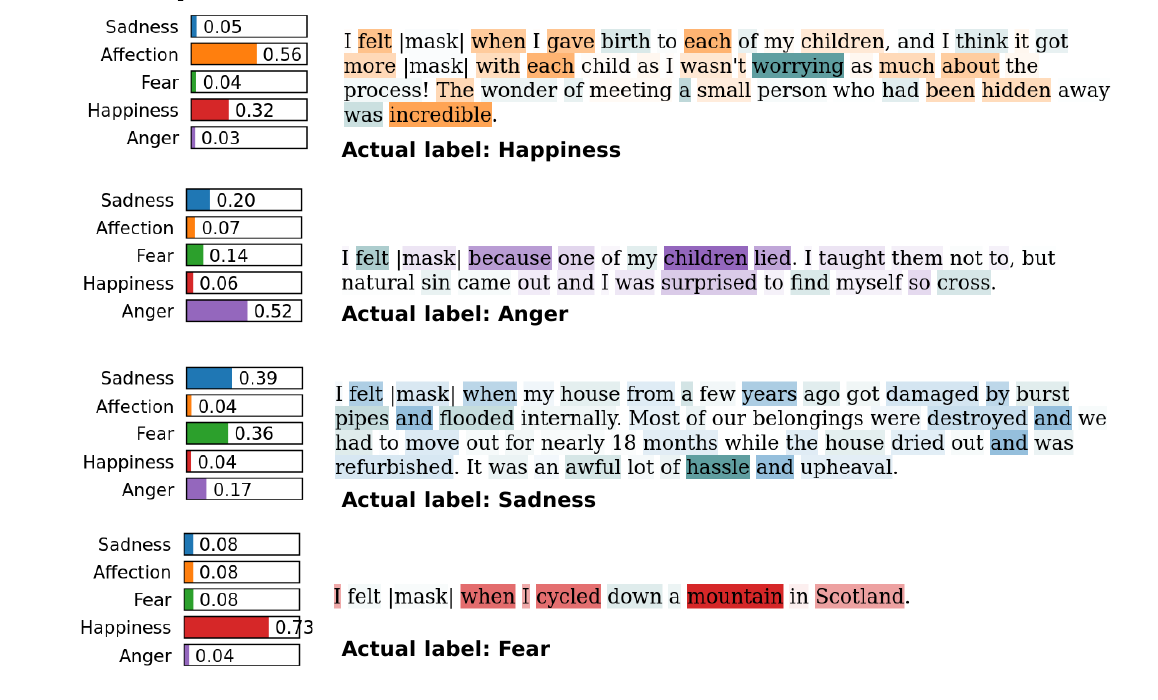} 
    \caption{LIME explanations showing the feature importance for LEIA-base prediction on four examples taken from the enISEAR dataset. The mask token is $<mask>$, shown with vertical lines in the figure.
    }
    \label{fig:lime_explanation}
\end{figure*}

\paragraph{Feature attributions} We examine the salient features that contribute to the predictions made by LEIA-base on a set of examples from the enISEAR dataset. We apply the Local Interpretable Model-agnostic
Explanations (LIME) method for model interpretability~\cite{lime}, an attribution method for  identifying salient features as n-grams of the classified text.
Figure \ref{fig:lime_explanation} shows four examples, one for each class of emotions in the enISEAR test set. The first column shows the model confidence scores for each class supported by LEIA-base and the text is colored according to which words contribute to the prediction.

We observe that for the first example, the model incorrectly predicts Affection as the most likely label where the true label is Happiness, which is an error of a weaker kind since enISEAR does not have an Affection label and both emotions are close in terms of valence. The second highest class is Happiness and the prediction is positively based on words expressing high arousal and valence (e.g., "incredible") and negatively based on the word "worrying".
In the second example, the model also seems to use relevant words linked to each other (e.g., "children" and "lied") to make the correct prediction. 
The model correctly predicts Sadness for the third example building on negative words, including terms linked to property damage that caused an emotional loss.
We observe that the scores for fear and sadness are very close and much higher than for other classes. This seems plausible as the first sentence in this example could be a fearful situation.
The model prediction is Happiness in the fourth example instead of Fear, which was the true label. Even though the prediction relies on relevant features, the model seems to lack the commonsense knowledge that cycling down a mountain can be scary and not necessarily a pleasant experience.

The last two cases suggest that the emotion tag for some of the posts is used as the main medium to express the emotion, leaving the text to add other information. This is one of the limitations of using Vent as a training dataset, as labels are part of the communication and may sometimes be complementary or otherwise to the posts.

\section{Discussion}
We present LEIA, a language model in two sizes (LEIA-base  and LEIA-large), that leverages approaches for adapting pre-trained language models for emotion identification. We show that using an emotion lexicon with task-adaptive pre-training, in this case focusing on emotion words, is effective for improving model performance using BERTweet-base and BERTweet-large language models. 
LEIA generalizes beyond Vent posts as it shows better performance on texts written by users not included in its training data and future time periods. It achieves a balanced performance across emotion labels despite their imbalance in training data and this performance is also seen on out-of-domain texts for the considered emotions except for Fear. These results are in part possible thanks to focusing on a small set of emotions suggested by psychological research, as classifying the larger set of mood labels in Vent \cite{alvarez-gonzalez-etal-2021-uncovering-limits} is a substantially harder task we did not tackle here.
Also, the Vent dataset, which despite being generated on a platform not as large as common ones in research, e.g. Twitter and Reddit, has a sufficiently large scale that enables the models to learn a broader range of emotional expressions. 

The performance of LEIA-base is comparable to LEIA-large across tests in our benchmark with one notable exception: LEIA-large is substantially better for the enISEAR dataset. This dataset is especially important given the psychological methodology used to generate it, which allows us to compare the results of machine learning methods with self-reported labels in a controlled setup. LEIA's performance in enISEAR is especially high, reaching F1 of 70 for LEIA-base and 79 for LEIA-large, showing a high level of psychological validity, especially when compared to other methods in the benchmark that achieve at most 55. LIWC generally achieves low F1 in all tests except SemEval, which grants two notes. First, SemEval was generated by searching tweets with emotion-bearing terms, easing the task of LIWC when classifying emotions based on similar word lists. Second, LIWC was not designed as an emotion classification method at the scope of a social media post, but as a more general text analysis method that should be applied to longer texts and not necessarily for classification. We added LIWC as a contrast with common methods applied in the field, but our comparison overstretches the applications for which LIWC was designed.

\paragraph{Limitations} While we show that our proposed models are effective, our experiments span two model sizes with the same architecture. Future research should conduct experiments on other pre-training approaches beyond masking as well as more efficient training techniques. In addition, we rely mostly on hyperparameter settings in the literature and optimizing them could lead to better performance. However, this is computationally expensive and there might be unfavorable trade-offs between model performance and resources. Another limitation is our focus mainly on English posts,  providing no evidence here of the potential of this approach for other languages. 
Furthermore, we study five emotion labels guided by psychological research, but several competing representations models for emotion are available. Humans are able to classify a larger number of basic emotions and can also quantify emotions in dimensional spaces, two open areas that can be explored with more nuanced labeling schemes.
While self-annotated datasets have the potential to become the new gold standard beyond crowdworkers, the Vent dataset is still produced with visible mood labels rather than private reports that do not constitute part of online communication. This is still closer to general emotion expression than automatic labeling with emoji or hashtags, but models like LEIA-base or LEIA-large can be substantially improved with psychological methods like experience sampling with validated psychological scales \cite{elayan2020stresscapes}.

\paragraph{Broader impact and ethical considerations}
This work shares the same ethical concerns with other emotion recognition systems as highlighted in \cite{mohammad2022ethics}. Emotion detection models should be used responsibly and special care should be taken when they are applied in new scenarios, not only because of their possible lower performance but also due to possible different privacy expectations with respect to emotions. We must note that we have no way of estimating the demographic diversity of Vent users and it is very likely that the model misses idiosyncrasies of emotional expression in minority groups and in cultures not represented in the dataset. We acknowledge that we only consider one type of model evaluation focusing on accuracy while there are several aspects such as bias, fairness, and robustness that should be considered before a model is used in practice, especially when guiding any decision-making.

\section{Conclusion}
LEIA is an emotion detection method that achieves a balanced performance across emotions and generalizes across posts, users, and time. It shows satisfactory performance in out-of-domain tests, especially when compared to self-annotated texts produced with psychological methods. Beyond our validations, the language models within LEIA can be used as pre-training resources for future applications that employ annotated data in other domains, for example for tweets in particular contexts.

We named LEIA after Princess Leia from Star Wars, following the tradition of emotion method names set out by LIWC \cite{chung2012linguistic} (pronounced Luke, as in Luke Skywalker), and VADER \cite{hutto2014vader} (as in Darth Vader). These three methods have a similar purpose but very different approaches that align with concurrent developments in text analysis.  
We published openly our models in HuggingFace (\url{https://huggingface.co/LEIA}) including both the classifier LEIA-base (LEIA-large) and the corresponding emotion-aware language model with the hope that they can be used in future work in emotion detection from text.

\section*{Funding}
The research leading to these results received funding from the Vienna Science and Technology Fund (WWTF) [10.47379/VRG16005]. DG and SA acknowledge funding from the ERC Advanced Grant PRODEMINFO (101020961).

\section*{Author's contributions}
DG, HM, and ADN designed research.
LM and NH contributed to model development.
SA designed and trained the final model and performed all analyses.
DG and SA wrote the first manuscript draft.
All authors read and approved the final manuscript.

\newpage
\section*{Appendix}
\label{additional_results}
\paragraph{Comparison of intermediate models} Recall that we average the parameters of two model variants. A BERTweet-base model fine-tuned on the labled Vent training set and a BERTweet-base with eMLM pre-training before fine-tuning to derive LEIA-base. We follow the same approach for BERTweet-large to derive LEIA-large. Here, we  compare the performance of each of these models on in-domain and OOD test sets. Results in Table~\ref{tab:intermediate_indomain_comp} show that LEIA-base and LEIA-large have the highest average rank across in-domain test sets. We also see a similar pattern in Table~\ref{tab:intermediate_ood_comp} for the OOD datasets. This suggests that taking an average of model parameters contribute to the generalization of LEIA-base and LEIA-large. In addition, the eMLM pre-training step is also effective given that the models with this pre-training step rank second across both in-domain and out-of-domain test sets.

\begin{table*}[!hbtp]
    \centering
    \caption{Comparison of performance of intermediate models on in-domain test sets.}
    \resizebox{\textwidth}{!}{
        \begin{tabular}{lllll}
        {} &                user &            temporal &              random & Average rank \\ \hline
        BERTweet-base       &  72.82[72.72,72.91] &  72.85[72.77,72.95] &   72.92[72.81,73.0] & 3.00\\
        BERTweet-base+eMLM  &  72.87[72.79,72.97] &   72.91[72.82,73.0] &   73.0[72.92,73.09] & 2.00\\
        LEIA-base   &  72.92[72.82,73.02] &  73.03[72.95,73.11] &  73.02[72.94,73.12] & 1.00\\ \hline 
        BERTweet-large      &  73.03[72.94,73.11] &   73.0[72.91,73.12] &  73.19[73.09,73.28] & 3.00\\
        BERTweet-large+eMLM &  73.19[73.11,73.28] &  73.14[73.05,73.23] &  73.37[73.27,73.46] & 2.00\\
        LEIA-large  &  73.37[73.28,73.46] &  73.43[73.34,73.53] &  73.57[73.48,73.66] & 1.00 \\ \hline
        \end{tabular}
    }
    
    \label{tab:intermediate_indomain_comp}
\end{table*}

\begin{table*}[!hbtp]
    \centering
    \caption{Performance comparison of intermediate models on out-of-domain datasets}
    \resizebox{\textwidth}{!}{
        \begin{tabular}{lllllll}
        {} &           Universal Joy &          GoEmotions &                 TEC &             SemEval &           enISEAR & Average rank\\ \hline 
        BERTweet-base       &   52.40[47.01,57.99] &  47.16[44.74,49.61] &   43.68[41.90,45.41] &   69.70[67.14,72.25] &   67.23[63.30,71.06] & 2.60\\
        BERTweet-base+eMLM  &  52.07[46.85,57.61] &  46.26[43.89,48.66] &  42.43[40.64,44.16] &   70.45[67.90,72.97] &  74.79[71.24,78.22] & 2.00\\
        LEIA-base   &  54.18[48.79,59.88] &  46.31[43.98,48.72] &  43.87[42.05,45.61] &  71.68[69.18,74.19] &  70.37[66.63,74.01] & 1.40\\ \hline
        BERTweet-large      &  51.87[46.77,57.32] &  45.51[43.14,47.94] &  44.08[42.28,45.85] &   68.75[66.20,71.31] &  80.49[77.18,83.72] & 2.40\\
        BERTweet-large+eMLM &  52.28[47.01,57.79] &  46.94[44.58,49.31] &  43.78[41.98,45.57] &  69.02[66.47,71.55] &  80.09[76.77,83.37] & 2.00\\
        LEIA-large  &  54.17[48.68,59.84] &  45.75[43.45,48.09] &  44.12[42.34,45.89] &  70.04[67.48,72.52] &  79.94[76.69,83.14] & 1.60\\ \hline
        \end{tabular}
    }

\label{tab:intermediate_ood_comp}
\end{table*}

\end{document}